# Geometry of Language

Loe Feijs
*Eindhoven University of Technology*

**Abstract.** In this article, we present a fresh perspective on language, combining ideas from various sources, but mixed in a new synthesis. As in the minimalist program, the question is whether we can formulate an elegant formalism, a universal grammar or a mechanism which explains significant aspects of the human faculty of language, which in turn can be considered a natural disposition for the evolution and deployment of the diverse human languages. We describe such a mechanism, which differs from existing logical and grammatical approaches by its geometric nature. Our main contribution is to explore the assumption that sentence recognition takes place by forming chains of tokens representing words, followed by matching these chains with pre-existing chains representing grammatical word orders. The aligned chains of tokens give rise to two- and three-dimensional complexes. The resulting model gives an alternative presentation for subtle rules, traditionally formalized using categorial grammar.

## 1. INTRODUCTION

The quest for a kind of fundamental understanding of language is an important intellectual undertaking. Already in antiquity, scholars understood that language is not random, but that each language is governed by a precise set of rules. Grammar books have a long history, for example, Panini's Sanskrit grammar dates to the 6th to 5th century BCE. In the early 19th century, Grimm and others discovered that language evolution is also not random but follows precise rules. Although a variety of formalisms for describing languages have been proposed, some of the fundamental questions have not been adequately answered. Key questions are: is there a unique language instinct built-in to the human brain, and if so, how does it work? Is there a formalism which is minimal, yet capable of describing each human language?

The introduction of production-style grammar rules by Noam Chomsky was an important step forward (Chomsky 1957). Later Chomsky introduced the concept of UG, universal grammar (Chomsky 1965). The idea is that, although different human languages have different grammars, there could a set of structural rules, innate to humans. In the Minimalist program, the hypothesis was that there could be a kind of minimal kernel which supports the quick learning of any language offered to a child. But it was hard to identify this minimalist meta-grammar and, in more recent years Chomsky holds it that perhaps the only faculty of language in the narrow sense (FLN) is a recursive computational system (Hauser et al. 2002).

Another significant contribution was the introduction of type theory in the form of categorial grammars (Moortgat 1988, Van Benthem 1991). The study of language analysis has been mostly driven forward by efforts to let computers do analysis work and a large portion of present-day grammatical theory is influenced by or geared towards, the languages of computers and mathematical logic.

In this article, while being informed by production-rules grammars and categorial grammars, we take a slightly different approach, which is new to the best of our knowledge. We acknowledge the inspiration from early writings on "denksoep" (Dutch, i.e. thinking soup) by N.G. de Bruijn (1996), next to inspiration from DNA computing (Adleman1994). We work from omnipresent phenomena of human language, trying to find simple rules about two- and three-dimensional complexes which explain the language phenomena. The author misses the knowledge and the tools to search or check for the actual molecules and neuron networks in our brains. The result is a kind of model which explains aspects of language analysis. Focus is on the syntactic aspects of language, and although the model contains clues and handles for dealing with meaning, we leave semantics as an option for future research.

## 2. SUBJECT-VERB-OBJECT SENTENCES

In this section, we limit ourselves to simple sentences of the subject-verb and subject-verb-object types. These correspond to the word order in English phrases and in Dutch non-complementizer phrases. Extensions to more complicated examples will be provided later, in Sections 3−7.

The most obvious phenomenon of human language is the sequential nature of its form (not necessarily of its meaning). Words are produced one after the other, both in speaking and in writing. Speech understanding thus starts from the same spoken word order, and so does reading.

The **first assumption** we adopt in our model is that the words are converted into compact tokens, each token representing the spoken occurrence of a word. Here we write them as strings, such as "cows", "eat", and "grass". Later we also depict them as coloured balls, which is convenient for modelling and showing 3D complexes. Additional tokens represent grammatical categories, such as *NP* (noun phrase), $V_1$ (non-transitive verb), $V_2$ (transitive verb), *Adj* (adjective) and *S* (sentence).

The **second assumption** is that tokens are connected into linear chains such that the token order in the chain corresponds to the word order of the spoken or written sentence. We write them from left to right and add hyphens to connect them, as in "cows"—"eat"—"grass". In addition to such chains of words, it is also possible to have chains which contain grammatical categories, for example, a mixed chain "cows"—$V_2$—"grass'", or a completely abstract chain *NP*—$V_2$—*NP*.

The **third assumption** is that every now and then when a chain is heard and is confirmed to belong to a certain grammatical category, a token representing that category is added. We write it at the right-hand side end of the chain after an arrow sign. For example, "cows"—"eat"—"grass"→*S* which represents the knowledge that *cows eat grass* is a sentence, or "cows"→*NP*, classifying one specific word, or *NP*—$V_2$—*NP* → *S* which represents the abstract knowledge of the word order in this type of sentence. This lastly added category token is said to be the *conclusion*. Knowledge of language is represented by a set of such chains. Everyone has a different set of chains, depending on what he or she has heard and memorized (it is even possible that there are many areas or vessels within one person with different such sets, but for the time being we work with one set).

In a grammar based on production rules, such as proposed by Chomsky (1957), similar knowledge is coded by production rules, in particular the rule S ⇒ *NP* $V_2$ *NP* for the non-terminal *S* and the rules for the terminal symbols *NP* ⇒ "cows", *NP* ⇒ "grass", and of course $V_2$ ⇒ "eat".

The **fourth assumption** is that during processing a new sentence, the newly heard chain is matched against the pre-existing chains. The chains are aligned in the same direction (head-to-head and end-to-end). This matching works such that equal tokens attach to each other in pairs and turn the aligned chains into more complex structures we call *complexes*. The complex is *complete* if it forms a closed mesh with precisely one conclusion dangling. This conclusion is what is picked up by the environment of the matching processes (perhaps the dangling conclusion drops of the complex and triggers the action-upon-recognition). So if an *S* is the only loose end of the complex, the owner of this brain knows he or she recognized a sentence.

We show the above four assumptions in action. Let the pre-existing chains be "cows"→*NP*, "grass"→*NP*, "eat"→$V_2$, and *NP*—$V_2$—*NP*→*S*. Next, let us assume that in this context the chain "cows"—"eat"—"grass" enters, after which the complex shown in Figure 1 forms.

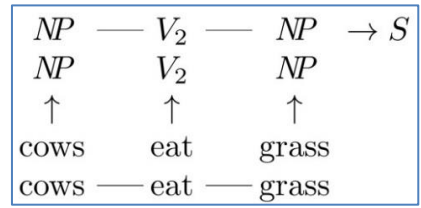

Figure 1: Complex formation of *cows* eat grass.

The fresh chain is usually written underneath and the complex formation proceeds upwards. We allow for one exception to the rule that the chains must be aligned parallel: the conclusions of chains such as "cows"→*NP* are allowed to be bent upwards. As the example shows, the entire complex becomes a kind of commuting diagram with two inner loops and an outer loop, and the *S* is the only dangling end. This *S* triggers the environment, in other words, "cows"—"eat"—"grass" represents a correct sentence. Of course, we need not use all pre-existing chains. On the other hand, for each pre-existing chain we assume that enough copies are available (otherwise we could run into trouble with recursion). We show the ingredients which are sufficient for successful recognition of the sentence in Figure 2.

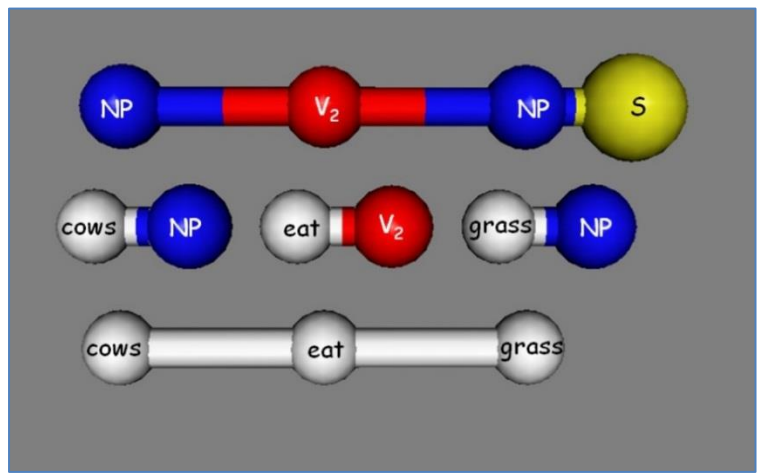

Figure 2: Ingredients for *cows eat grass* to be recognized as *S*.

We show the complexes as drawn by the molecular visualization program Molekel version 5.4.0.8 (ugovaretto.github.io), which was designed primarily for showing molecules, but whose stick-and-ball pictures also work for the complexes studied here. In Molekel we can rotate them in 3D space and see them in perspective, but in this paper, we show flattened renderings. We use white balls for raw word tokens, blue for nouns or noun phrases (*NP*), red for verbs (here $V_2$), and a big yellow ball for the *S*. We use long sticks for the horizontal links, corresponding to the temporal ordering, from left to right. We use shorter sticks for the connection between a chain and its conclusion (either horizontal or bent upwards), and no stick when equal tokens are attached to each other (vertically), see Figure 3.

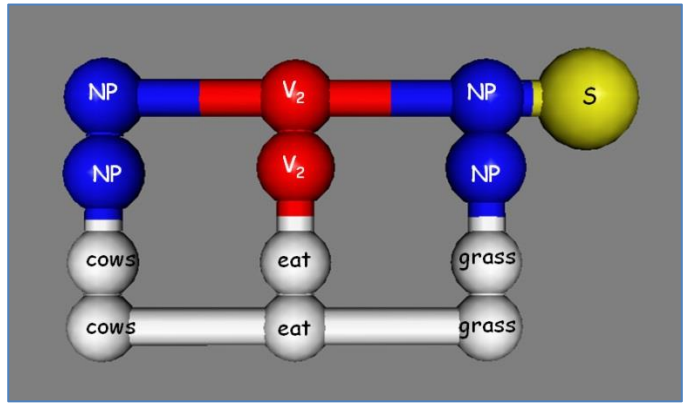

Figure 3: *Cows eat grass* recognised as *S*.

## 3. LEAVING THE 2D PLANE

Now we address grammatical sub-constructs, for example, adjectives. As a classical production rule, an adjective would be expressed by a rule such as *NP* ⇒ *Adj NP* for the non-terminal *NP* and rules for terminal symbols *Adj* ⇒ "big", *Adj* ⇒ "brown", and so on. In our own model, we assume corresponding pre-existing chains "big"→*Adj*, "brown"→*Adj*, and *Adj*—*NP*→*NP*. Now assume that in this context the chain "brown"—"cows"—"eat"—"grass" enters. If we analyze this in a two-dimensional format we must allow the conclusions of chains not only to be bent upward but also being stretched, see Figure 4.

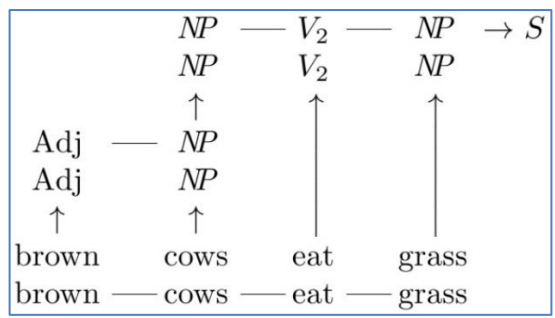

Figure 4: *Brown cows eat grass* recognized as *S*.

But if we allow the same construction to happen in 3D space, a little bending and rotating suffices, see Figure 5. We show adjectives as grey balls. Although one could think that this freedom allows for any mix of ingredients to yield an *S*, this is not the case. Note that there is the rule that alignment follows the original temporal ordering, in other words, each chain has a beginning and an end. In the stick-and-ball figures, this ordering is not shown, except for the fact that we use renderings where this ordering roughly follows the horizontal *x*-direction. In other words, the beginning of a chain is to the left, and the end is to the right. We could add arrow signs in the sticks, but for the time being, the rendering condition works fine (Figure 5).

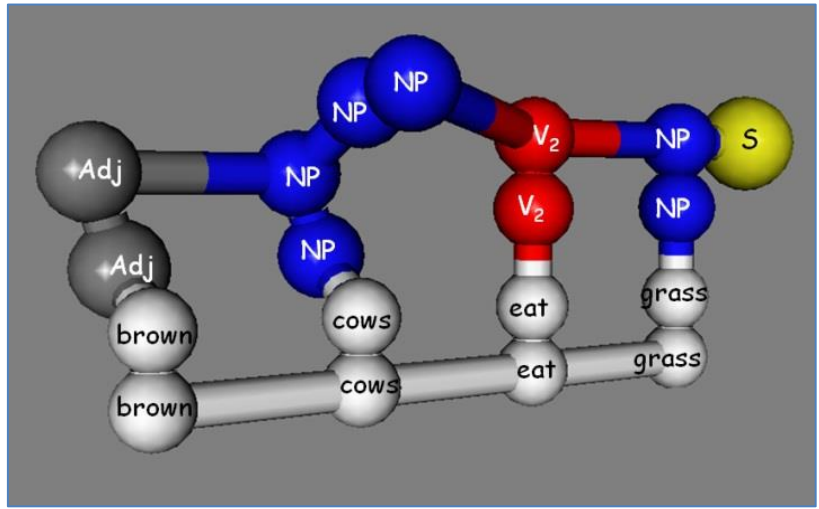

Figure 5: *Brown cows eat grass* recognized as *S*.

## 4. MIXED CHAINS

It is not necessary that all words are classified into grammatical categories. There can be consolidated chains which still have fragments at a word level, such as idiomatic phrases or phrases not yet well understood. In the next example, consider sentences of the form "it's"—"okay"—"to"... . In the model, let the context contain pre-existing chains including "eat"→$V_2$, "grass"→*NP*, and "it's"—"okay"—"to"—$V_2$—*NP*→*S*. Assume that in this context the chain "it's"—"okay"—"to"—"eat"—"grass" enters. In 2D notation, we get the complex of Figure 6.

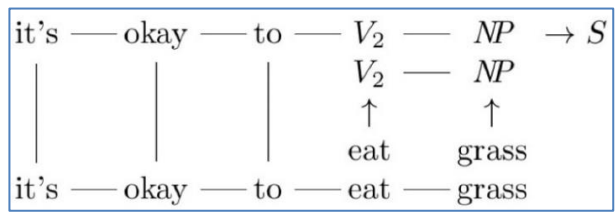

Figure 6: *It's okay to eat grass* recognized as *S*.

The complex is complete since it forms a closed mesh with precisely one dangling conclusion, viz. *S*. This is recognized as a sentence. Presenting the geometry of the complex in 3D, we see that the problem of the long sticks between the words disappears. Equal tokens attach in pairs easily, see Figure 7.

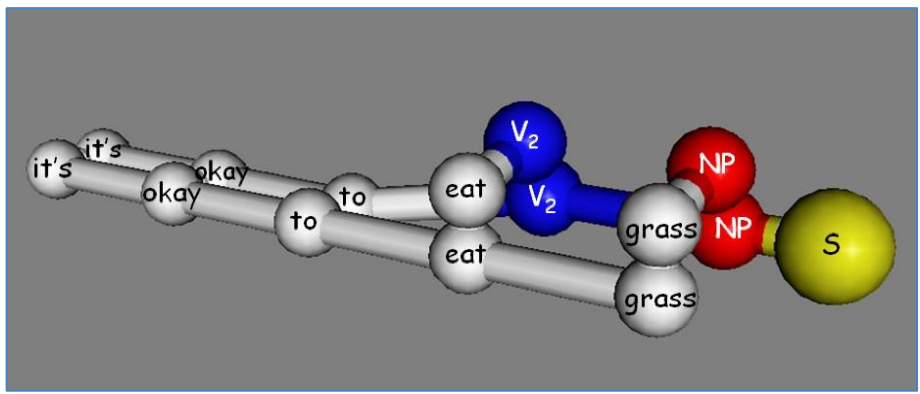

Figure 7: *It's okay to eat grass* recognized as *S*.

So far, the reader could object that the new model only was used to replicate analyses which are well within reach of traditional production rule grammars. But now we present examples where the model works with specific chains with mostly real words, not only abstract grammatical categories and where we show how a kind of inductive learning takes place, just by building complexes.

In the next example, the model works with a mix of grammatical categories and actual words. In the example, there is one word which does not have a grammatical category yet, but the recognition succeeds by means of analogy. Assume the following chain has been heard before and is known to be a sentence: *pigs eat beans*. Let "cows", "pigs", "eat" and "hate" be known, but assume "beans" is not. Let the context thus contain pre-existing chains "cows"→*NP*, "pigs"→*NP*, "eat"→$V_2$, "hate"→$V_2$ and finally "pigs"—"eat"—"beans"→*S*. Now let us assume that, in this context, the chain "cows"—"hate"—"beans" comes in. Even without assigning *NP* to "beans", it is induced by analogy that "beans" works as an *NP* (it does in one context the same it does in the other context).

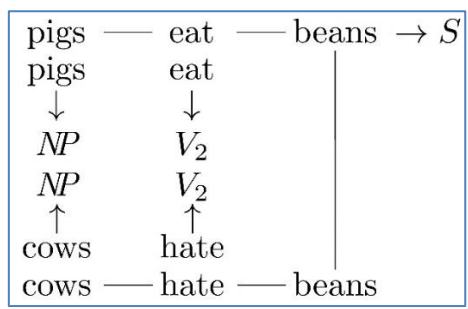

Figure 8: *Cows hate beans* recognized as *S*.

In 3D representation, the chains can be aligned without stretched sticks. Again, here is one dangling end, an *S*, which therefore is the conclusion of the complex shown in Figure 9.

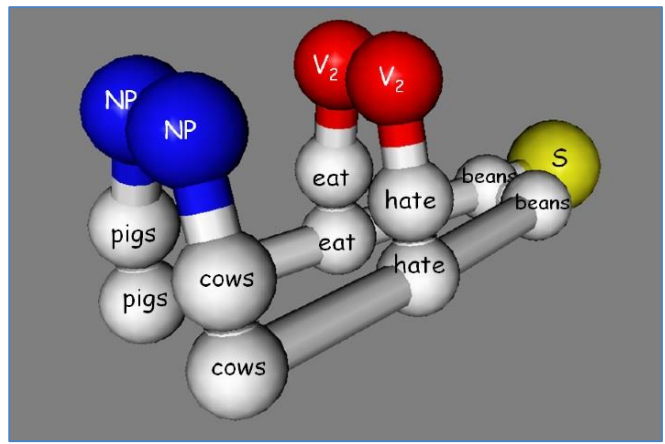

Figure 9: *Cows hate beans* recognised as *S*.

In the next example, which is also a kind of analogy, the geometry of the complex is even more exciting. Assume the following chains are heard and are confirmed as being sentences: *birds fly*, *bats fly*, and *birds sing*. Let this knowledge be memorized, which in the model means that the context contains pre-existing chains "birds"—"fly"→*S*, "bats"—"fly"→*S*, and "birds"—"sing"→*S*. Next, let us assume that, in this context, the chain "bats"—"sing" enters.

Even without the introduction of formal grammatical categories such as *NP* and $V_1$ one can induce by analogy that "bats" can do (grammatically) what "birds" do and so if "birds" followed by "sing" makes an *S* then "bats" followed by "sing" make an *S* too. It is hard to do this in the 2D plane; a best effort is shown in Figure 10: *Bats sing* recognised as *S* by analogy.Figure 10.

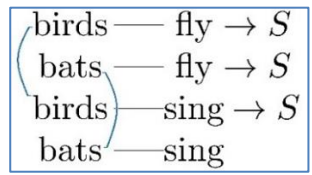

Figure 10: *Bats sing* recognised as *S* by analogy.

However, in 3D this works perfectly. Note that the *S* conclusions attach to each other as well. The complex is complete, see Figure 11. There is one dangling end, an *S*, which therefore is the conclusion of the complex. In other words, "bats"—"sing" is a correct sentence (syntactically, here we do not care whether bats can really sing).

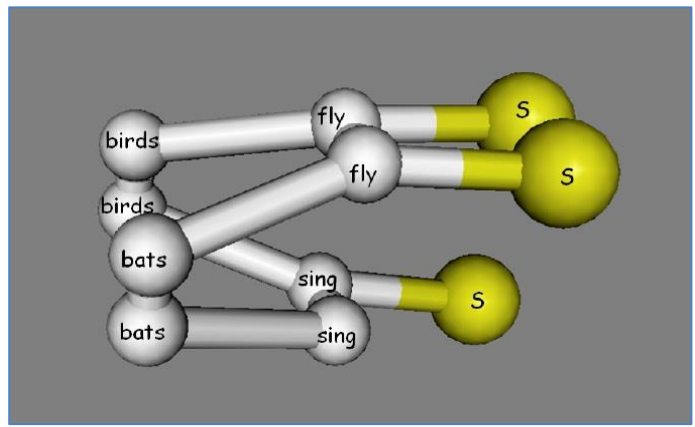

Figure 11: *Bats sing* recognized as *S* by analogy.

## 5. CASES

Next, we address a linguistic phenomenon which is not elegantly captured by the formalism of traditional Chomsky-style production rule grammars. It is the phenomenon of specific noun phrases which are meant for a subject position or a direct object position, but not for both. Whereas the sentence *he loves Mary* is correct, it is not okay to use "he" in the direct object position, as in *he loves he* (wrong). Instead, the special form "him" has to be used for that. In older languages such as Latin, Sanskrit and old versions of Dutch and German the phenomenon is part of an elaborate system of *cases*. The form "he" is said to be the first case and "him" the fourth case, or nominative case and accusative case, respectively.

If we try to do this with a production rule grammar we could have $NP_1 \Rightarrow$ "he" and $NP_4 \Rightarrow$ "him". Next to the common rule for sentence $S \Rightarrow NP\ V_2\ NP$ one has to add two special versions $S \Rightarrow NP_1\ V_2\ NP$ and $S \Rightarrow NP\ V_2\ NP_4$. But it is ugly that *he loves him* still is not accepted unless yet another special rule $S \Rightarrow NP_1\ V_2\ NP_4$ is added.

Moortgat used grammatical categories which are similar to the types used in mathematical type theory (Feys and Curry 1958, Lambek 1958) and thus solved this problem. The resulting grammars are known as categorial grammars. For example, a grammatical category such as $NP\backslash S$ is meant for words which leftwardly look for something of type *NP* and then result in something of type *S*. This is precisely what a intransitive verb does: when there is a noun phrase left to it, the result is a sentence. Similarly, $S/NP$ is looking to its right for something of type *NP*. Bracketing occurs inside types, so for example, $(NP\backslash S)/NP$ is a type too (it is the type of transitive verbs). A colon is used to indicate typing, so for example "eat" : $V_2$ means that "eat" has type $V_2$. We present the most important rules used by Moortgat next. Each rule comes in two versions.

- (R1) if $f : B/A$ and $a : A$ then $f\,a : B$      (right version)
- (R1) if $f : A\backslash B$ and $a : A$ then $a\,f : B$      (left version)
- (R2) if $f : A/B$ and $g : B/C$ then $f\,g : A/C$      (right version)
- (R2) if $g : C\backslash B$ and $f : B\backslash A$ then $g\,f : C\backslash A$      (left version)
- (R3) if $f : (C\backslash A)/B$ then also $f : C\backslash(A/B)$      (right version)
- (R3) if $f : C\backslash(A/B)$ then also $f : (C\backslash A)/B$      (left version)
- (R4) if $a : X$ then also $a : B/(A\backslash B)$      (right version)
- (R4) if $a : X$ then also $a : (B/A)\backslash B$      (left version)

Rule R1 is called *application*, R2 *composition*, R3 *associativity* and R4 is *lifting*. A sequence is a correct sentence if there is a derivation which assigns type *S* to it.

We show a few examples of derivations in Moortgat-style categorial grammar. The first example is *birds sing*. Let "birds" : *NP* and "sing" : $V_1$ where $V_1$ is considered an abbreviation of $NP\backslash S$. Figure 12 shows that *birds sing* has type *S*.

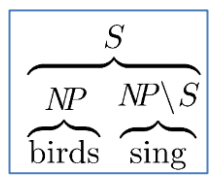

Figure 12: *Birds sing* derived as having type *S*.

Now we present an example with an intransitive verb and an adjective, viz. *brown cows eat grass*. Of course "cows" : *NP*. Let "eat" : $V_2$ where $V_2$ is nothing but an abbreviation of $(NP\backslash S)/NP$. Let "brown" : *NP/NP*, so the adjective looks for an *NP* to its right and if it finds it, the composition of the adjective and noun phrase has type *NP* again. The type $NP_1$ is defined as $S/(NP\backslash S)$, whereas the type $NP_4$ can be defined as $(S/NP)\backslash S$. Figure 13 shows one possible derivation.

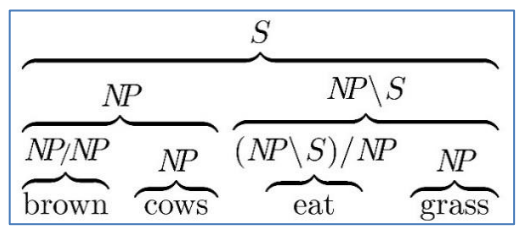

Figure 13: *Brown cows eat grass* derived as having type *S*, Moortgat style.

In fact, this is not the only derivation, since by rule R3 the word "eat" also has type $NP\backslash(S/NP)$, so "eat" can be combined with "brown cows" first. These ambiguities, which turn out harmless, are typical of Moortgat-style categorial grammars.

Now we are ready to address the "he" and "him" issue, which is the third example. The type $NP_1$ was already defined as $S/(NP\backslash S)$, whereas the type $NP_4$ can be defined as $(S/NP)\backslash S$. Let "he" : $NP_1$ and "him" : $NP_4$. The sentence *he loves Mary* is derived easily, as shown in Figure 14.

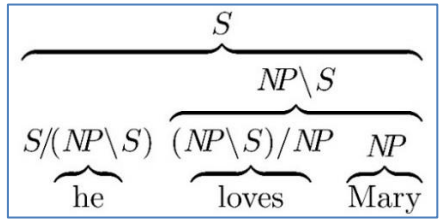

Figure 14: *He loves Mary* derived as having type *S*, Moortgat style.

In the same way, *Mary loves him* is derived easily. Now it is a remarkable property of the Moortgat-style categorial grammar that the recognition of *he loves him* comes for free. This is because of rule R2, composition. We show this in Figure 15, using the right version of rule R2.

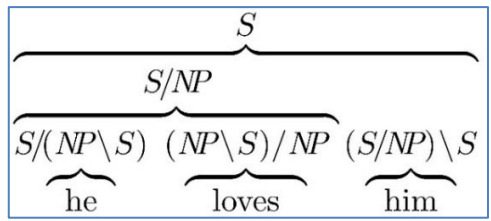

Figure 15: He loves him derived as having type as S, Moortgat style, right version of rule R2.

After this excursion into Moortgat's approach, we show how the same sentence is addressed using the model of 3D complexes. Assume the following chain has been heard before and is known to be a sentence: *pigs eat beans*. Let the context also contain pre-existing chains "he"→$NP_1$, "loves"→$V_2$, "him"→$NP_4$ which assign abstract tokens to specific words. Let the context also contain pre-existing chains $NP$—$V_2$—$NP$→$S$ describing transitive verbs, $NP_1$—$V_2$—$NP$→$S$ describing nominative case noun phrases, $NP$—$V_2$—$NP_4$→$S$ describing accusative case noun phrases. Just as in Moortgat's approach, no special provisions for the "he"-"him" combination have to be made, see Figure 16.

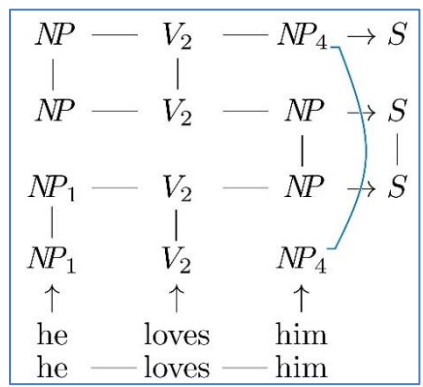

Figure 16: *He loves him* recognized as *S*.

In 3D presentation, the corresponding complex is shown in Figure 17. Note that this complex is somewhat similar to the Moortgat derivation in the sense that first the *he loves* fragment is processed and found to be something which combines with an $NP_4$ to its right in order to become a sentence.

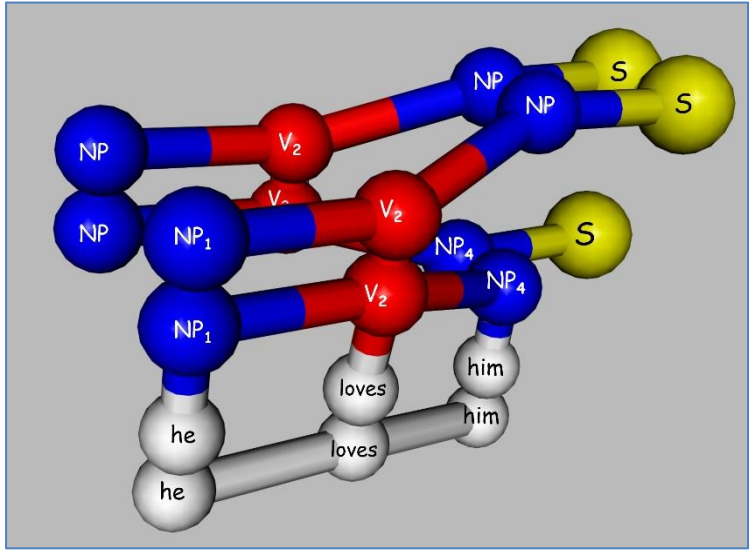

Figure 17: *He loves him* recognized as *S*, similar to the Moortgat style derivation.

The first Moortgat derivation of *he loves him*, as shown above, is only one of several possible derivations, as was to be expected. The next derivation has the same result but this time "loves" combines with "him" first. In order to make this happen, the brackets in the type of "loves" have to be moved, but because of rule R3, associativity, that is no problem, as shown in Figure 18.

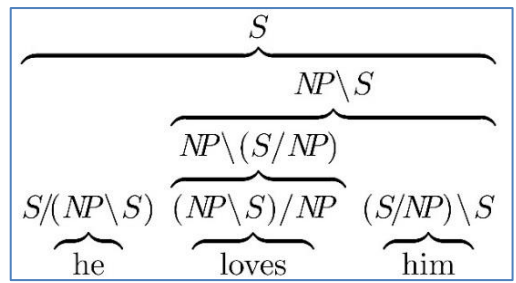

Figure 18: *He loves him* derived as having type *S* (alternative derivation).

Just like there are multiple Moortgat derivations, there are multiple complexes possible. In Figure 19, we show another one.

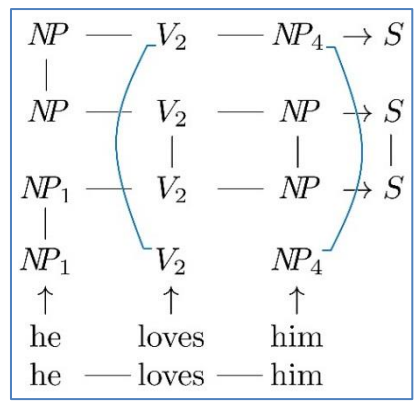

Figure 19: *He loves him* recognized as *S*, different complex.

In 3D presentation, the same complex is shown in Figure 20. It is remarkable that the same tetraeder-like geometric configuration of Figure 11 reappears in Figure 17 and (somewhat modified) in Figure 20.

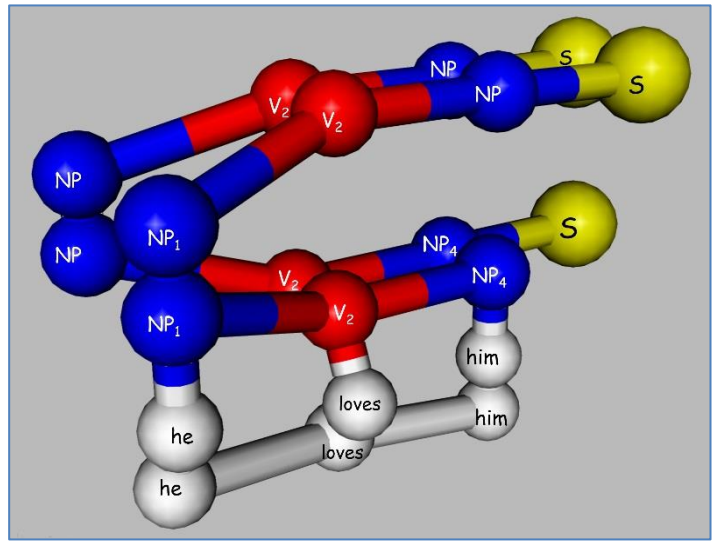

Figure 20: *He loves him* recognized as *S* (alternative complex).

## 6. INFLEXION AND CONGRUENCE

Besides word order, there are other interesting grammatical features, notably inflexion, cases, and congruence. Many of our contemporary languages seem to be losing those features, but in many classical languages, they are even more important than word order. However, if there something like universal grammar or if there is some truth in the ideas behind the minimalist program, then some aspects inflexion, cases, and congruence must be related to our innate language capacity too.

Flexion means that the words are modified depending on the grammatical role they have in the sentence. Verbs have variable vowels to reflect tense (for example *lesen*/*lasen* is present/past tense in German, to read) or elaborate systems of endings. Nouns have different endings to reflect cases (*rosa*/*rosae*/*rosam* for nominative/genitive and dative/accusative in Latin, rose). Also adjective have cases (nominative/genitive/dative/accusative). It is most remarkable that the sounds used for the noun cases are the same or almost the same as the corresponding adjective cases. Congruence means that noun and verb if connected, have the same case and number. Therefore, the words that are connected sound the same. If they are connected by syntax, they are connected in meaning. For example, *aurea fibula* in Latin means the golden brooch in nominative whereas *auream fibulam* means golden brooch in accusative. This observation, that similar-sounding things belong together, was, in fact, a decisive inspiration for our matching mechanism: equal tokens attach in pairs (the fourth assumption in Section 2).

What has our model to offer for modeling inflexion and congruence? We investigate this by analyzing a sentence in Latin from the book Aeneid (IV.139) written by Vergil (70BC-19BC), viz. *aurea purpuream subnectit fibula vestem*. This type of sentence is called a golden line. The nominatives, characterized by their "a" ending belong together. Similarly, the accusatives, characterized by their "m" ending belong together. Word order hardly matters in Latin, hence the meaning: *a golden brooch fastens her purple dress*.

In Figure **1** this is modelled by assuming that two "a" tokens combine into another "a" and similarly for the "b"-type tokens. The $V_2$ verb *subnectit* needs an "a"-type token and a "b" token. We have two types of noun phrases now and therefore we gave the tokens two different colours: blue for the "a" (nominative), purple for the "m" accusative.

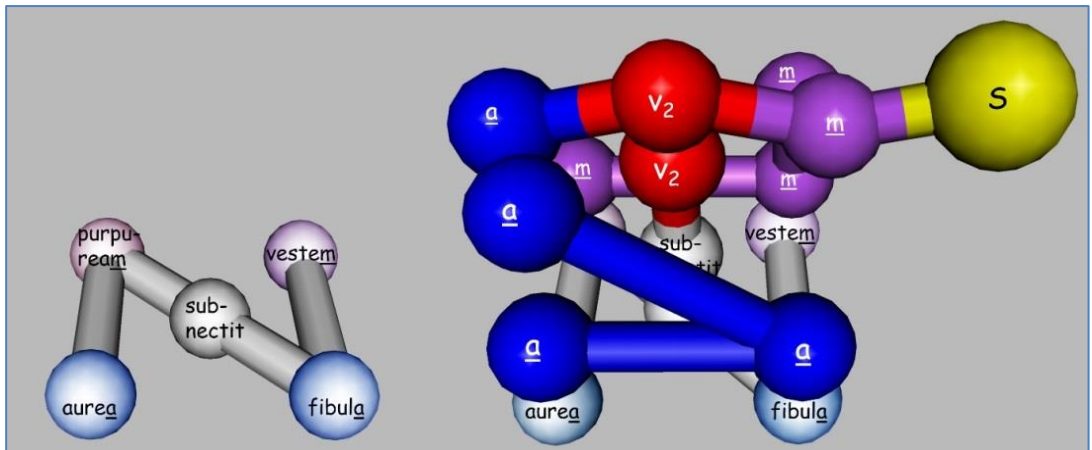

Figure 21: *Aurea purpuream subnectit fibula vestem* (left) and idem recognized as *S* (right).

We conjecture that the phenomenon that our language faculty connects similar-sounding tokens, is related to rhyme. With rhyme we refer to poetry, such as by Baudelaire (1821-1867): *L'un t'éclaire avec son ar<u>deur</u> / L'autre en toi met son deuil, Na<u>ture</u> ! / Ce qui dit à l'un : Sépul<u>ture</u> ! / Dit à l'autre : Vie et splen<u>deur</u> !* In Germanic languages, with their emphasis in the beginning of the words, it is also about similar beginnings; *<u>h</u>eerlijk <u>h</u>elder Heineken* (Dutch advertisement for beer). Modern languages, notably English, have less inflexion and thus modern grammar theory has focused on word order, and our computer languages, inspired by production rule grammars (Naur et al. 1976) and type theory (Milner 1978), have no inflexion whatsoever. But as Vergilius' example and the poems show, the human innate language capacity is not about word order only.

## 7. RECURSION

Recursion is the ability to place one component inside another component of the same kind. Quoting from (DeWitt 2013): "In linguistics, the core application of recursion is phrase embedding. Chomsky posits an operation, *unbounded Merge*, that recursively merges words to create larger phrases."

In this section, we show a possible analysis of the (Dutch) sentence *geiten haten kinderen die lawaai maken* (goats hate children who make noise). The relative pronoun *die* marks the subordinate sentence, but also plays the role of a noun phrase in the latter sentence. In Dutch, the word order is subject-verb-object, but in this complementizer phrase, as in other subordinate sentences, there is a reverse work order, viz. subject-object-verb. In a production rule grammar, the folowing rules would describe the generation of this sentence: $S \Rightarrow NP\ V_2\ NP$ and $CP \Rightarrow$ "die" $NP\ V_2$ and finally $NP \Rightarrow NP\ CP$ where *geiten*, *kinderen* en *lawaai* are *NP* and *haten*, *maken* are $V_2$. The complementizer phrase (*CP*) works as a postfix adjective. We show the complex in Figure 22.

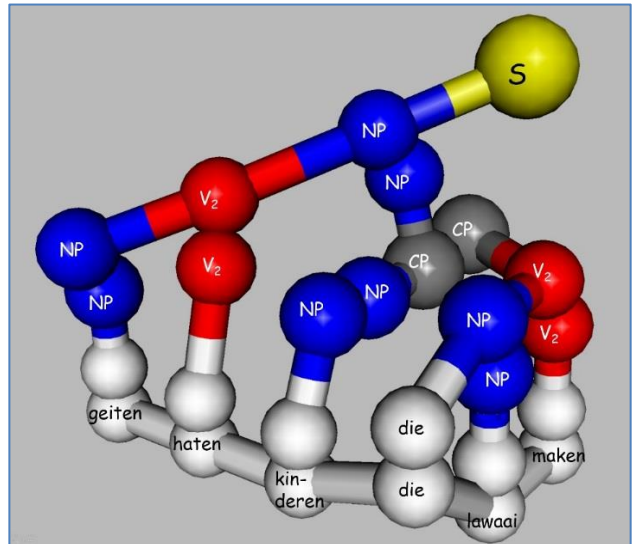

Figure 22: Complementizer phrase *die lawaai maken* recognized as *CP* and *sentence geiten haten kinderen die lawaai maken* recognized as *S*.

The token complex in Figure 22 is rather complicated. Therefore we like to point out another possibility, namely that there are several containers of tokens, where the recognized output of one container is fed into the input of another. This is sketched in Figure 23, where the leftmost container triggers upon recognition of a *CP* and the rightmost container triggers upon recognition of an *S* (Sentence).

Although we can only speculate on the implementation of such mechanisms in the human brain, we like to point out that there is a potential advantage of the mechanism with multiple containers. The number of pre-existing chains in each container can thus be limited, and a kind of work-division takes place. The pre-existing chains represent knowledge of the language which has been acquired before, such as "lawaai"→*NP* and "maken"→$V_2$ and "die"—*NP*—$V_2$→*CP*, in the leftmost container. All kinds of other and incomplete tokens may be present in the containers, but in Figure 23, only the ingredients needed for the successful recognition are shown. Even many types of containers are possible, corresponding to the various grammatical categories.

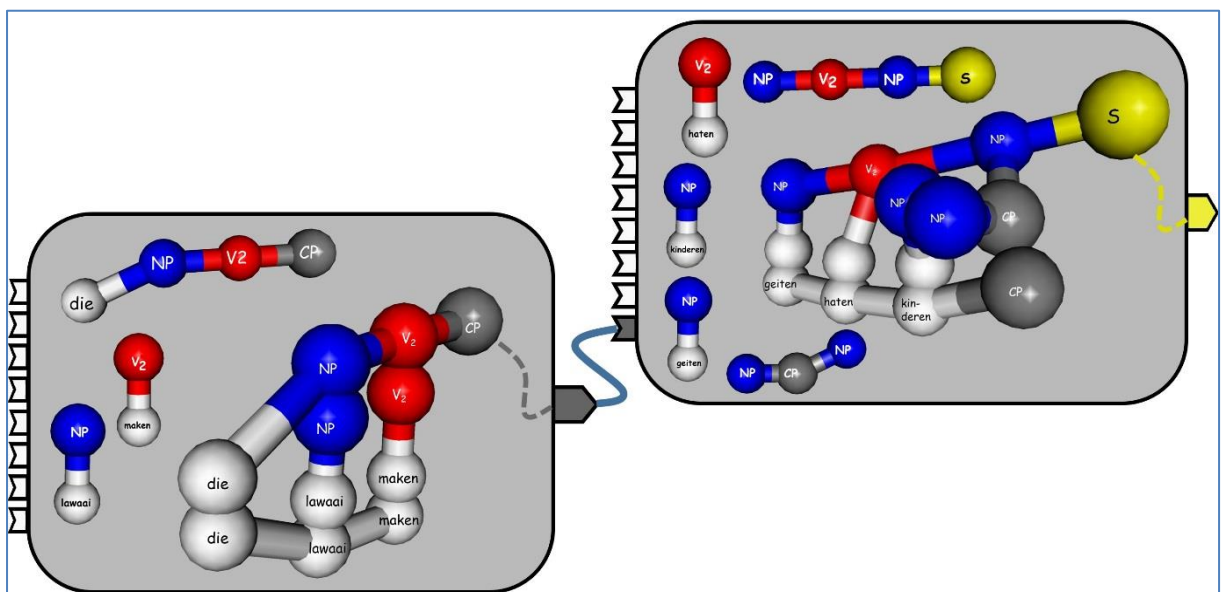

Figure 23: *Die lawaai maken* recognized as Complementizer Phrase and *geiten haten kinderen CP* recognized as *S* in two distinct token containers.

# 8. RELATED WORK

The goal to find a simple yet powerful mechanism or set of rules that characterise the human language faculty has been the driving force behind Chomsky's multi-decade quest for grammatical theory (Chomsky 1995), (Higginbotham 1998). Chomsky's *minimalist program* embodies the search for a minimal set of principles valid for all languages, a kind of set of possibilities that the children have innate and use when learning any natural language offered to them. The underlying idea is that the human language faculty has a kind of optimal and computationally efficient design. Similarly, Pinker's writings on the *language instinct* (Pinker1994) defend the thesis that language is innate and that humans have a kind of common "universal grammar".

The author started his explorations around 1995 and was inspired by De Bruijn's "denksoep" (thinking soup), e.g. found in (De Bruijn1996). De Bruijn writes: In the simplest case such reactions are about a compound *A* and a compound *B* giving rise to a compound *C*, expressed by $A + B \to C$. This is not necessarily an ordinary chemical equation where *C* is entirely composed out of *A* and *B*: it is also possible that *C* is being built from fragments floating around in the soup and that this synthesis is brought about with the cooperation of *A* and *B* [...] There may be many such reactions $A_i + B_i \to C_i$ where the index *i* can take al large number of values. (end quote).

We found the first version of the he-loves-him complex (cf. Figure 17) around 2000 but only later it was picked up again, and we found a way to write things down in a more satisfactory way.

We also noticed Maiya Sershen (2004) mentioning the concept of *sentence molecules* in Speculative Grammarian, but we were not able to find other proper sources. Unlike the chains and complexes proposed here, Sershen says the sentence molecules are (quote) large protein molecules in neural cells which almost precisely mimic the binary-branched tree structures already familiar to linguists worldwide. (end quote).

Cedric Boeckx (2017) discusses chunking devices in the brain, one of which is the so-called "global workspace", another is the fronto-thalamo-basal ganglia loop, originally dedicated to motor sequences.

Our work also resembles the work, entitled *geometry of language* too, by Glyn Morrill (1997). Just as in the present article, Morrill's work is inspired by Lambek Calculus (Lambek 1958) and the way Moortgat applies it to natural language. As in other so-called resource logics, Morrill's formalism gets rid of the sequential ordering of traditional sequent calculi and considers the assumptions in the context as a set or a multiset rather than a sequence. To make sure that the word order is not completely lost Morrill works with a non-planarity condition for the network that shows how the given words fulfil the needs of the various antecedent-succedent pairs. Example (37) from (Moririll 1997) gives the idea, see Figure 24.

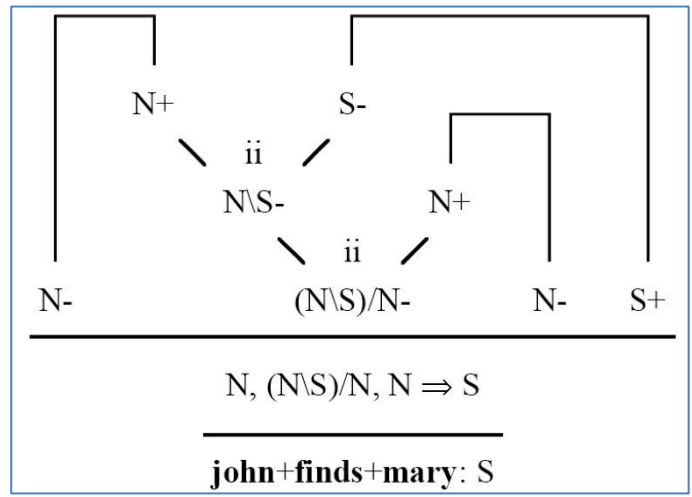

Figure 24: *John finds Mary* recognised as *S* in Morrill's formalism.

Thus Morrill sticks to 2D representation. The idea that the network could be a 3D complex is not made explicit (nor are the rules about closing the complex as embodied in our assumptions 1−4 of Section 1). Of course, Morrill is right to preserve word order, but in our approach, this is achieved by the sequential order in the various chains and the condition that they are aligned more or less alongside each other. Another distinction is that Morrill's tokens still carry markings for antecedent and succedent, e.g. $N^+$ and $S^-$, respectively, whereas in our fourth assumption we just let equal tokens attach to each other, without such markings. Actually, the idea that equals attach to each other is also inspired by observations about rhyme and congruence, as discussed in Section 6.

In the meantime, Moortgat (2002) has worked along another line, integrating syntax and semantics exploiting the Curry/Howard/De Bruijn propositions-as-types analogy. The idea is that an analysis of types (grammatical categories) is done and at the same time a lambda-calculus term construction around the original lexical tokens takes place. Thus logical aspects of meaning are embodied in the resulting the lambda-calculus term, somewhat as in Montague's proper treatment of quantifiers (Montague 1974).

## 9. CONCLUDING REMARKS

Clearly, the work presented here is exploratory, and somewhat outside the lines of contemporary mainstream linguistic research. We consider the analysis of the examples in Section 4 (analogical reasoning) and Section 5 (cases) particularly elegant and we conclude that a meaningful geometric view on language is possible. We hope it is a valuable contribution to the quest for an understanding of what is the essence of the human language facility. Many questions remain, such as:

- Is it possible that there are actual biological mechanisms in our brains, which can process chains and form complexes? Is it possible that the tokens are receptor-type molecules forming chains, somewhat like mRNA and that neuron cells play the role of containers (as suggested in Figure 23)? Or is there another neural computation mechanism which simulates the token-chains, in which case various brain areas play the role of containers?
- What are the limits to bending chains and stretching connections to let the complexes recognise good sentences, yet avoiding that "anything goes"? Can we fully formalise the proposed model of tokens, chains and complexes and compare its expressive power in a mathematical sense to known grammar formalisms such as production-rule based grammars, X-bar theory or categorial grammars?

How to validate the model? The best approach seems to build a simulator in software and feed such simulator with input from books, mail-boxes or chat-boxes by way of training. This approach is a new project, an option for later research.

**REFERENCES**

Adleman, L. (1994). Molecular computing of solutions to combinatorial problems, *Science*, Vol 266, pp.1021−1024 1994).

Van Benthem, J., (1991). *Language in action, categories, lambdas and dynamic logic*, North-Holland (1991).

Boeckx, C. (2017). A conjecture about the neural basis of recursion in light of descent with modification. *Journal of Neurolinguistics*, 43, 193−198.

De Bruijn, N.G. (1996). Can people think? *Journal of consciousness studies*, 3, No. 5−6, pp.425−47, 1996.

Chomsky, N (1957). *Syntactic Structures*, The Hague/Paris: Mouton de Gruyter.

Chomsky, Noam, 1965, Aspects of the Theory of Syntax, Cambridge, Massachusetts: MIT Press.

Chomsky, N. (1995). *The Minimalist Program*. (Current Studies in Linguistics 28.) Cambridge, MA: MIT Press. pp. 420.

Curry, H.B.; Feys, R. (1958). Combinatory Logic. Vol. I. Amsterdam: North Holland.

DeWitt, I. Language evolution and recursive thought. *Front. Psychol*., 30 October 2013. https://doi.org/10.3389/fpsyg.2013.00812

Hauser, M. D., Chomsky, N., & Fitch, W. T. (2002). The faculty of language: what is it, who has it, and how did it evolve? Science, 298(5598), 1569−1579.

Higginbotham, J. (1998). Visions and revisions: A critical notice of Noam Chomsky's The Minimalist Program. *Mind & language*, 13(2), 215−224.

Lambek, J. (1958). The mathematics of sentence structure, *American Mathematical Monthly* (65) 154−170.

Milner, R. (1978). A theory of type polymorphism in programming. *Journal of computer and system sciences*, *17*(3), 348−375.

Montague, R. (1974). The proper treatment of quantification in ordinary english. In Richard Thomason, editor, Formal Philosophy. *Selected Papers of Richard Montague*, pages 24−271. Yale University Press, New Haven and London.

Moortgat, M.J. (1988). *Categorial investigations, logical and linguistic aspects of Lambek calculus*, Proefschrift, Universiteit van Amsterdam (1988).

Moortgat, M.J. (2002). Categorial grammar and formal semantics. In: *Encyclopedia of cognitive science* #231, Nature Publishing Group, MacMillan (2002).

Morrill, G. (1997). Geometry of Language, *Report de Recerca* LSI-97-45-R, Departament de Llenguatges i Sistemes Informàtics, Universitat Politècnica de Catalunya.

Naur, P., Backus, J. W., Bauer, F. L., Green, J., Katz, C., McCarthy, J., & Perlis, A. J. (1976). *Revised report on the algorithmic language Algol 60*. Association for Computing Machinery.

Pinker, S. (1994). *The Language Instinct: The New Science of Language and Mind*, Penguin.

Sershen, M. (2004). The Biological Basis Of Universal Grammar − Speculative Grammarian*, specgram.com, Volume CXLIX*, Number 2, April 2004.